\documentclass{article}

\usepackage{arxiv}

\usepackage[utf8]{inputenc} % allow utf-8 input
\usepackage[T1]{fontenc}    % use 8-bit T1 fonts
\usepackage{hyperref}       % hyperlinks
\usepackage{url}            % simple URL typesetting
\usepackage{booktabs}       % professional-quality tables
\usepackage{amsfonts}       % blackboard math symbols
\usepackage{nicefrac}       % compact symbols for 1/2, etc.
\usepackage{microtype}      % microtypography
\usepackage{lipsum}
\usepackage{graphicx}
\graphicspath{ {./images/} }

\title{\LARGE \bf
DSFEC: Efficient and Deployable Deep Radar Object Detection
}

\author{
 Gayathri Dandugula \\
  Robert Bosch GmbH\\
  Corporate Research, India\\
  \texttt{gayathri.dandugula@in.bosch.com} \\
   \And
 Santhosh Boddana \\
  Robert Bosch GmbH\\
  Corporate Research, India \\
  \texttt{santhosh.boddana@in.bosch.com} \\
  \And
 Sudesh Mirashi \\
  Robert Bosch GmbH\\
  Corporate Research, India \\
  \texttt{sudeshganapati.mirashi@in.bosch.com} \\
}

\begin{document}
\maketitle
\begin{abstract}
Deploying radar object detection models on resource-constrained edge devices like the Raspberry Pi poses significant challenges due to the large size of the model and the limited computational power and the memory of the Pi. In this work, we explore the efficiency of Depthwise Separable Convolutions in radar object detection networks and integrate them into our model. Additionally, we introduce a novel Feature Enhancement and Compression (FEC) module to the PointPillars feature encoder to further improve the model performance. With these innovations, we propose the DSFEC-L model and its two versions, which outperform the baseline (23.9 mAP of Car class, 20.72 GFLOPs) on nuScenes dataset: 1). An efficient DSFEC-M model with a 14.6\% performance improvement and a 60\% reduction in GFLOPs. 2). A deployable DSFEC-S model with a 3.76\% performance improvement and a remarkable 78.5\% reduction in GFLOPs. Despite marginal performance gains, our deployable model achieves an impressive 74.5\% reduction in runtime on the Raspberry Pi compared to the baseline. 
\end{abstract}

% keywords can be removed

\section{Introduction}
In recent years, Radar systems have become indispensable in enhancing the perception abilities of autonomous vehicles, especially in challenging weather conditions. The Radar data plays a critical role in enabling accurate object detection, collision avoidance and adaptive cruise control, thus ensuring safe navigation and consistent performance across various driving scenarios. However, current state-of-the-art radar-based algorithms require significant GPU computing power and frequently encounter challenges in achieving real-time object detection. Yet, achieving real-time processing is crucial for autonomous vehicles to enable timely decision-making and responses to dynamic changes in the environment, ensuring safe and efficient navigation by processing sensor data, detecting obstacles and executing control commands with minimal latency. 

State-of-the-art object detection models, designed for Image-based \cite{liu2016ssd}\cite{he2017mask} and Lidar-based \cite{yang2018pixor}\cite{zhou2018voxelnet} methods have demonstrated high accuracy in various performance metrics. Initially, radar-based methods lagged behind due to the sparsity of the data compared to other sensor modalities.However, recent studies have improved our understanding of how radar can be beneficial for autonomous vehicles. \cite{scheiner2021object} provides a concise overview of the radar-based object detection methods. Recent studies have primarily focused on enhancing feature extractors rather than backbone architectures, often utilizing similar backbones\cite{yang2018pixor}\cite{qi2017pointnet}. Moreover, the emphasis has been on improving accuracy rather than deployability. In our work, we are exploring modifications to the backbone of \cite{yang2018pixor}\cite{ulrich2022improved} along with the feature extractor to emphasize the significance of real-time processing in autonomous vehicles. Our goal is to develop deployable models suitable for edge devices while maintaining high accuracy.
The significant contributions of our work are as follows:
\begin{itemize}
    \item Introduction of the FEC (Feature Enhancement and Compression) module to the feature extraction process, facilitating learning and mitigating memory bottlenecks at the initial stages of the network.
    \item We investigated the efficacy of depthwise separable convolutions and integrated them into the backbone of the radar-based object detection model \cite{ulrich2022improved}. This integration is aimed to enhance the model's efficiency, enhancing performance while minimizing latency.
    \item Designing a deployable version of the proposed model specifically tailored for ARM-based Raspberry Pi.
\end{itemize}

\section{RELATED WORK}

\subsection{Feature Encoder for Point Clouds}

Sensor data, unlike images, often arrives in irregular formats like point clouds, where each point signifies a measurement from the environment. Analyzing such data requires specialized techniques such as Point-based and Grid-based architectures to extract meaningful information effectively due to its irregular nature. Point-based architectures  \cite{bansal2020pointillism},\cite{qi2017pointnet}, \cite{qi2017pointnet++},\cite{schumann2019scene},\cite{danzer20192d} utilize raw point cloud data to extract features directly. PointNet\cite{qi2017pointnet}, a foundational work in this field, processes point clouds without requiring any pre-processing and PointNet++\cite{qi2017pointnet++} builds upon this by hierarchically grouping points and extracting features to improve performance. 

In grid-based approaches, point clouds are initially transformed into either a 2D bird's eye view (BEV) or a 3D voxel grid using hand-crafted or learned feature encoders. Hand-crafted feature extraction methods \cite{nobis2021radar},\cite{dreher2020radar},\cite{meyer2019deep},\cite{nesti2023ultra} rely on manual algorithm design based on domain knowledge or heuristics. However, in complex data like point clouds, these methods have limitations to capture all intricate feature relationships. Conversely, learned feature encoders, such as PointPillars\cite{lang2019pointpillars} and VoxelNet\cite{zhou2018voxelnet}, employ CNNs or point cloud processing networks to automatically learn these relationships from 2D BEV or 3D voxel grid points, potentially uncovering nuances missed by hand-crafted approaches. VoxelNet\cite{zhou2018voxelnet} employs voxelization to transform point clouds into evenly distributed 3D voxels. It utilizes Stacked Voxel Feature Encoding (VFE) layers to extract features. On the other hand, the PointPillars\cite{lang2019pointpillars} utilizes a CNN pillar feature net to extract features from raw point clouds and efficiently converts them into a pseudo-image format. We chose the feature encoder from the PointPillars network for its effective pseudo-image representation in terms of memory and computation. We then implemented our innovative approach within this feature encoder.

\subsection{Object Detection for Radar point cloud}
Feature Pyramid Networks(FPNs)\cite{lin2017feature} were introduced as single-stage object detection models, utilizing a top-down architecture with lateral connections to construct a feature pyramid from a backbone network. This design addresses the challenge of object detection in images with varying scales. RetinaNet\cite{lin2017focal}, an image-based object detector, utilizes FPNs on top of ResNet\cite{he2016deep} to generate features at different pyramid levels. Following a similar single-stage dense object detection approach, Pixor\cite{yang2018pixor} extends these concepts to real-time LiDAR-based 3D object detection by adjusting input representation, network architecture and output parameterization. Several radar-based object detection methods\cite{ulrich2022improved}\cite{niederlohner2022self}\cite{kohler2023improved} have adopted the Pixor\cite{yang2018pixor} framework for their work with three-headed network to address the challenge of detecting objects at different scales.

All of these networks employ ResNet\cite{he2016deep}'s residual blocks as the fundamental building blocks for their CNN backbones. While these residual blocks are renowned for their high accuracy in computer vision tasks due to the innovative skip-connection mechanism in standard convolutional networks and demand higher computational resources and memory. Consequently, this results in heightened latency on edge devices. Moreover, as more of these blocks are added in each stage, the latency increases proportionately. To tackle this issue, we conducted an analysis of various convolution types, including Dilated\cite{yu2015multi}, Quantized\cite{wu2016quantized} and Depthwise Separable\cite{chollet2017xception} convolutions, implemented across diverse applications. Particularly, the Depthwise Separable Convolution, introduced in XceptionNet\cite{chollet2017xception}, substitutes a standard convolution with a depthwise and a pointwise convolution, rendering them nine times faster and suitable for building lightweight architectures like \cite{jang2023falcon}, \cite{kaddar2021divnet}, \cite{tan2019mixconv}, \cite{qin2018merging}, \cite{howard2017mobilenets} for mobile applications.

Furthermore, MobileNet\cite{howard2017mobilenets} and its subsequent versions \cite{sandler2018mobilenetv2}, \cite{howard2019searching} leveraged depthwise separable convolutions, demonstrating superior performance in the ImageNet classification challenge. Additionally, Object detection methods \cite{chiu2020mobilenet}, \cite{ding2023yolov4}, \cite{chang2021position} have adopted depthwise separable convolution-based backbones, showcasing improved latency and accuracy on edge devices. These findings prompted us to delve deeper into this approach, leading to the adoption of Depthwise Separable Convolution(DSConv) as a replacement for the default residual block in our work as outlined in \cite{ulrich2022improved}.

\section{METHOD}
\label{sec:method}
\begin{figure*}[h]
    \centering
    \includegraphics[width=0.8\textwidth]{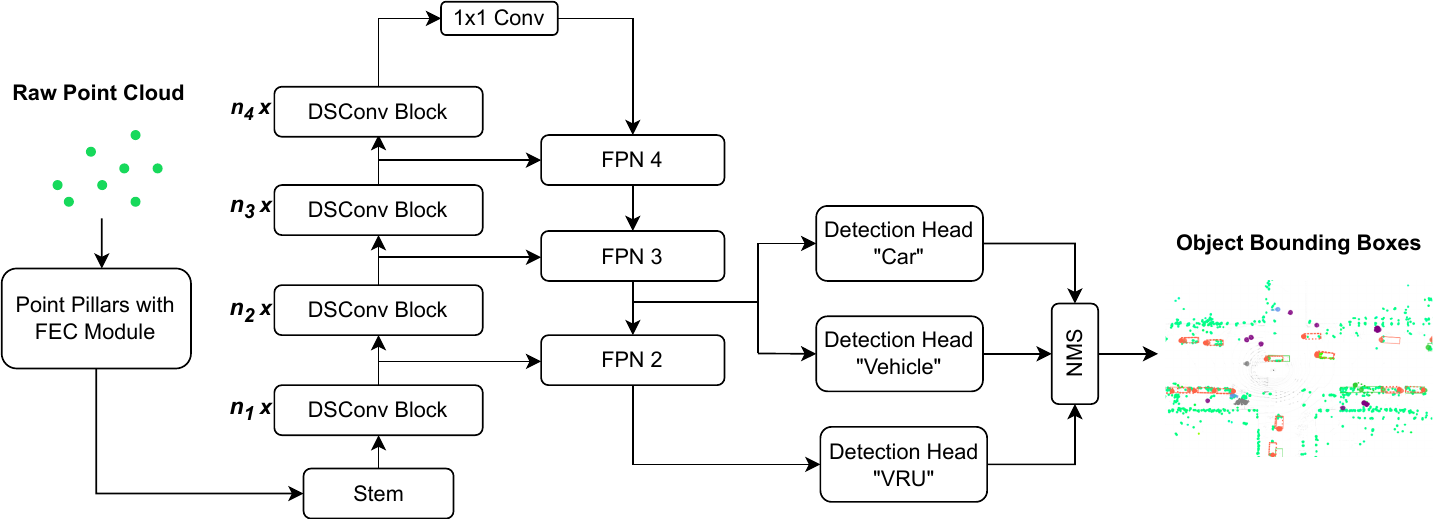}
    \caption{DSFEC: The radar object detection network takes raw point cloud as an input, applies feature encoder followed by backbone and detection heads. Non-maximum suppression (NMS) is then applied to suppress excess bounding boxes and produce final OBBs. The parameters $n_1$,$n_2$,$n_3$ and $n_4$ represent the number of blocks in four stages of the network, with different configurations designed for the proposed models.}
    \label{fig:architecture}
\end{figure*}

In this section, we present our innovative approach to enhancing the feature encoder and backbone of the network. We also discuss, how we designed our efficeint model and deployable model. An overview of the proposed radar object detection network is shown in Fig.\ref{fig:architecture}.

\subsection{FEC Module} 
The PointPillars \cite{lang2019pointpillars} explains the pseudo-images formation from 3D point clouds, which are then utilized within the 2D backbone. This approach involves several steps: discretizing the point cloud, stacking non-empty pillars, applying a 1x1 convolution and subsequently incorporating Batch Normalization and ReLU activation. However, due to the presence of a single convolution layer, the PointPillars network is constrained to either feature enhancement or compression, depending on the number of filters employed in that convolution. Emphasizing feature enhancement alone leads to an increase in the number of features, thereby causing the stem to become a runtime and memory bottleneck due to its utilization of 3x3 convolution on those resulting features. On the other hand, solely applying feature compression leads to the loss of information at the beginning of the network. This limitation hampers its ability to provide optimal feature encoding essential for the 2D backbone.

\begin{figure}
    \centering
    \includegraphics[width=0.45\textwidth]{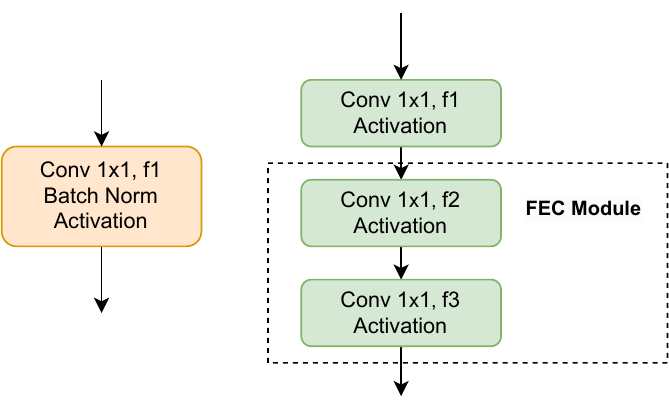}
    \caption{Left: Existing point wise convolution in PointPillars, Right: Proposed FEC module with 2 additional convolutions without Batch Normalization}
    \label{fig:fec}
\end{figure}

Given this analysis, we introduced a Feature Enhancement and Compression (FEC) module, which consists of two additional 1x1 convolution layers after the existing convolution layer, as illustrated in the Fig.\ref{fig:fec}. The first convolution is dedicated to feature enhancement by enabling the utilization of a greater number of filters(f2) to enhance feature representations. Subsequently, the second convolution, equipped with a limited number of filters(f3), focuses on feature compression. This approach allows the stem to process the compressed features efficiently with fewer filters, resulting in reduced runtime and memory footprint. Thus, our FEC module ensures both feature enhancement and compression simultaneously. To ensure this, the selection of filter numbers for these three layers (f1, f2, f3) should adhere to the following criteria: f1\(<\)f2, f2\(>\)f3, f1\(>=\)f3. Based on our experiments, we found that incorporating Batch Normalization after a 1x1 convolution in PointPillars led to a decline in performance, attributed to information loss at the initial stage of the network. Consequently, in our FEC module, we opted not to utilize batch normalization after convolution. Thus, our proposed network consists of three consecutive 1x1 convolution layers with ReLU activation as depicted in Fig.\ref{fig:fec}.

\subsection{Base Network Architecture}
We consider the network proposed in the paper\cite{ulrich2022improved} with the PointPillars feature encoder as our baseline and conducted benchmark experiments atop it. Inspired by the efficiency gains observed in cutting-edge methodologies\cite{chollet2017xception}\cite{ding2023yolov4}\cite{howard2017mobilenets} utilizing depthwise separable convolutions, We replaced the residual block utilized in \cite{ulrich2022improved} with the depthwise separable convolution (DSConv) block. This involves employing a combination of 3x3 depthwise convolution and 1x1 pointwise convolution to replace the standard 3x3 convolution. This replacement enhances the model's efficiency in terms of both accuracy and latency by reducing the number of parameters and FLOPs. In addition, adopting this approach enables the utilization of only a single convolution layer instead of two layers in the stem, similar to \cite{howard2017mobilenets}. Our DSConv block-based network, in conjunction with the FEC module discussed in Sec.\ref{sec:method}A, forms our new architecture, termed DSFEC, as illustrated in Fig. \ref{fig:architecture}. Our initial model DSFEC-L, adopts a configuration with the number of blocks per stage as $n_1$=3, $n_2$=6, $n_3$=6 and $n_4$=3, inspired by prior works \cite{yang2018pixor}\cite{ulrich2022improved}. This serves as the foundation for further model development and comparison.

\subsection{Designing DSFEC-M \& DSFEC-S Models}
From our ablation study Sec.\ref{sec:ablation}B focusing on DSFEC-L configuration, we observed that a network with less number of blocks could yield comparable performance with reduced runtime latency. This way, We designed an Efficient DSFEC-M Model by adjusting the number of blocks in the second and third stages to 3 and 2, respectively, resulting in $n_1$=3, $n_2$=3, $n_3$=2 and $n_4$=3. This architecture allowed us to a 60\% reduction in GFLOPs while maintaining higher mAP than the base architecture.

Although the DSFEC-M model demonstrated superior performance compared to the base architecture, it was not ideally suited for deployment on resource-constrained edge devices like the Raspberry Pi. Hence, we iteratively redesigned the model, leading to the creation of the DSFEC-S Model optimized for edge deployment. By further adjusting the number of blocks in each stage, we reduced runtime while preserving performance. The final configuration comprised $n_1$=1, $n_2$=1, $n_3$=3 and $n_4$=2, achieving an 80\% reduction in GFLOPs with a slight mAP improvement from 23.9 to 24.8 compared to the baseline.

Additionally, our experimentation highlighted the impact of activation functions on performance and runtime. While complex functions like Mish and Swish enhanced performance, they also introduced additional runtime overhead compared to simpler activations like ReLU and Leaky ReLU. To strike a balance between accuracy and efficiency, we propose employing different activation functions for various components of the backbone: ReLU for FEC module and the VRU head, Leaky ReLU for the stem and the backbone and Swish for the first two heads (Car \& Truck) in both our DSFEC-M and DSFEC-S models.

\section{EXPERIMENTS}

\subsection{Experimental setup}
All experiments were conducted using the TensorFlow framework. Our model was trained with a standard Adam Optimizer with a momentum of 0.9 and a weight decay of 0.01, over a duration of 30 epochs. The initial learning rate was fine-tuned to 0.00348 and updated using a Piecewise Constant Decay scheduler with 400 warm-up iterations, employing a batch size of 24. The experiments and comparisons presented were performed on the nuScenes public Radar object detection dataset. A grid-like structure of point cloud has been generated using X range of [0,80], Y range of [-40,40], Z range of [-2.5,2.5] and cell size of 0.5.

\begin{table*}[ht]
    \caption{Evaluation results of radar based object detection on nuScenes validation dataset. We benchmark the performance of our proposed DSFEC-L, DSFEC-M and DSFEC-S models for the \textit{CAR} class. Runtime mentioned is averaged over 6019 frames.}
    \centering
    \begin{tabular}{c c c c c cc cc}
        \hline
        \multicolumn{1}{c}{Model} & \multicolumn{1}{c}{mAP$\uparrow$} &  \multicolumn{1}{c}{AP@4$\uparrow$} & \multicolumn{1}{c}{GFLOPs$\downarrow$} & \multicolumn{1}{c}{Params$\downarrow$} &  \multicolumn{2}{c}{FPS$\uparrow$} & \multicolumn{2}{c}{Runtime(m.sec)$\downarrow$}\\
        \cline{6-9}
          &  &  &  &  &  V100 &  RasPi &  V100 &  RasPi \\
        \hline
         Baseline(Residual) & 23.9 & 39.3 &  20.72  & 2.63M & 44 &  0.78 & 22.51 & 1276 \\
         DSFEC-L &  27.4 & 42.8 & 18.96  & 2.37M  
         & 51 & 0.88 & 19.15 & 1136\\
         \textbf{DSFEC-M} & \textbf{27.4} & \textbf{41.9} & \textbf{8.29} & \textbf{0.58M} 
          & \textbf{55} & \textbf{1.5} & \textbf{18.04} & \textbf{668}\\
         \textbf{DSFEC-S} & \textbf{24.8} & \textbf{39.0} & \textbf{4.45} & \textbf{0.3M} 
          & \textbf{58} & \textbf{3.08} & \textbf{17.2} & \textbf{325} \\
         \hline
    \end{tabular}
    \label{tab:results}
\end{table*}

\subsection{Evaluation Metrics}
We evaluate our experiments using performance metrics defined by the nuScenes detection benchmark dataset. Although our network can detect Car, Truck, Pedestrian and Bicycle classes, we specifically emphasize the Average Precision(AP) at a distance of 4 meters and the overall mean AP(mAP) across all four distances ({0.5,1,2,4} meters) for the Car class to align with previous radar object detection literature. Additionally, we utilize the number of model parameters and GFLOPs to compare the sizes of various models. For benchmarking hardware performance metrics like Frames Per Second (FPS) and average runtime against the GPU v100, we opted for the Raspberry Pi 4 (8 GB RAM, 64-bit quad-core ARM Cortex-A72 processor) due to its superior processing power and lower power consumption compared to other hardware options. We serialized our models in Protobuf format for deployment and conducted evaluations on a dataset comprising 6019 test frames.
\subsection{Results}
Experimental results for various network configurations are summarized in Tab.\ref{tab:results}, focusing on performance metrics mentioned in above section. The Baseline model, which utilizes Residual Blocks, achieves an mAP of 23.9 but requires significant computational resources. Conversely, our proposed baseline model, DSFEC-L improves performance and efficiency over the Residual model as mentioned in Tab.\ref{tab:results}. The DSFEC-M model stands out for its optimal balance of high performance, retaining an mAP of 27.4 and an AP@4 of 41.9, while achieving a notable runtime reduction on Raspberry Pi. Meanwhile, the DSFEC-S model achieved good mAP and is mainly distinguished for its utmost deployability with the smallest model size, delivering a 74.5\% runtime reduction on Raspberry Pi. This model is especially suited for resource-constrained environments, emphasizing its practical application in real-world scenarios.

\begin{figure*}[h]
    \centering
    \includegraphics[width=1\textwidth]{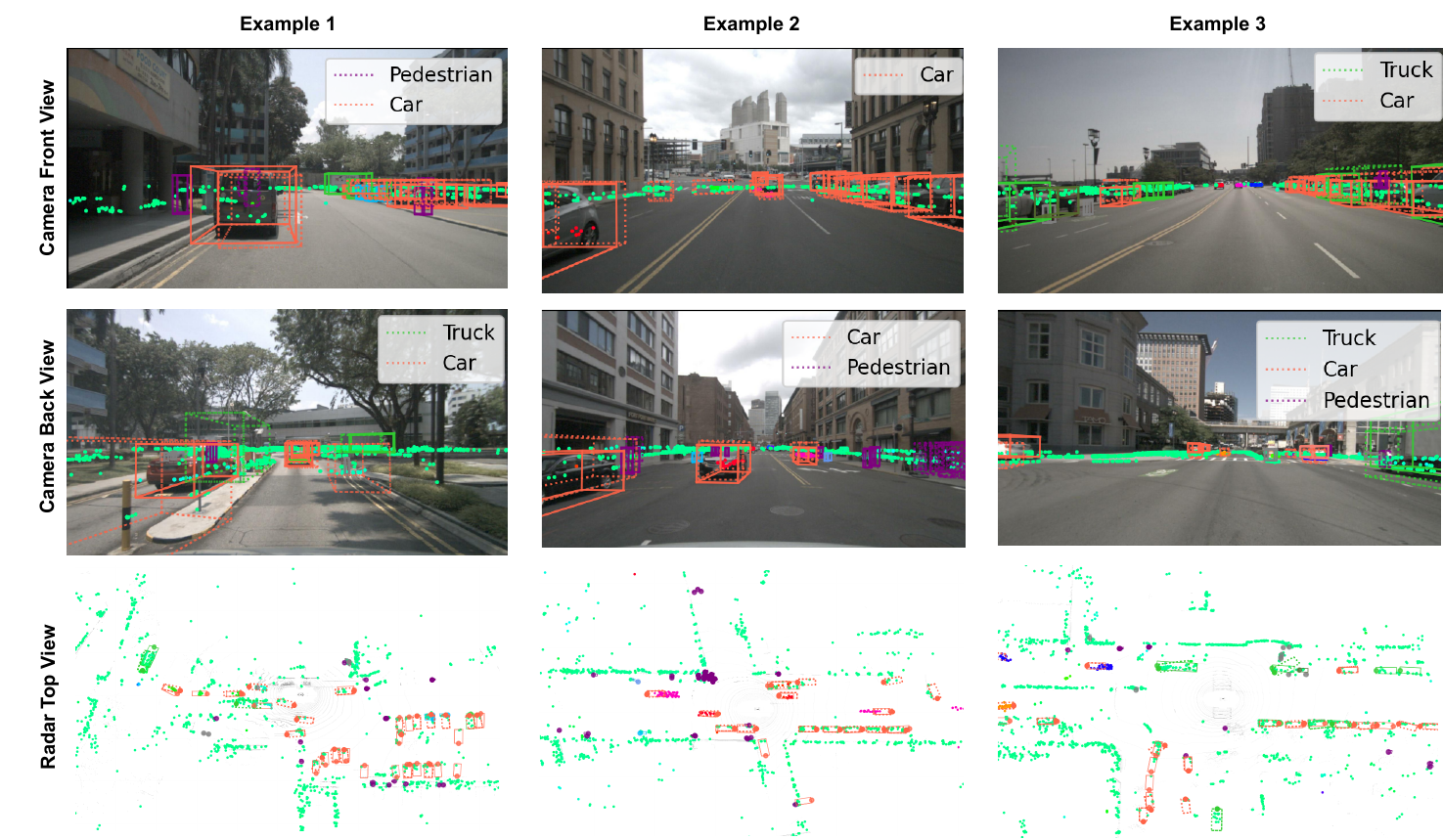}
    \caption{Qualitative results of our work on various samples from the nuScenes validation dataset. The predictions are visualized on radar top view, as well as front and back camera views. Here dotted lines indicates predicted boxes and solid lines indicates ground truth boxes, Different classes are distinguished by different colors.}
    \label{fig:results}
\end{figure*}

\section{Ablation Studies}
\label{sec:ablation}
In this section, we aim to validate the impact of each design
choice and hyperparameter setting of \cite{yang2018pixor}\cite{ulrich2022improved}.
\subsection{Stem Filters Analysis} We found that processing high-resolution input from PointPillars with the default stem (utilizing standard convolution with a 3x3 kernel and 32 filters) was causing runtime and memory bottlenecks. To address this issue, we utilized a configuration of 32, 128 and 12 filters in the FEC module, resulting in only 12 filters being required in the stem. Our experiments, as demonstrated in Tab.\ref{tab:stem}, revealed that reducing the stem filters to 12 resulted in a reduction of 0.72 GFLOPs, runtime by 20 m.sec and halved the memory usage to 2.35 MB while maintaining mAP, highlighting the effectiveness of the introduced FEC module and it's configuration. The highlighted row for filters = 12 underscores its superiority in balancing mAP, speed and resource consumption.

\begin{table}
  \caption{Ablation study of different filters for stem used in backbone of the network}
  \centering
  \begin{tabular}{ c | c | c | c | c}
    \hline
    Filters & mAP & GFLOPs & Runtime(m.sec) & Memory(MB) \\
    \hline
    32  & 27.4 & 1.06  & 33 & 4.32\\
    16  & 27.3 & 0.48 & 22 & 2.74\\
    \textbf{12}  & \textbf{26.5} & \textbf{0.34}  & \textbf{13} & \textbf{2.35}  \\
    8  & 25.4 & 0.19 & 9 & 1.96\\
    \hline
  \end{tabular}
  \label{tab:stem}
\end{table}

\subsection{Blocks per Stage Analysis} We noticed that the blocks in the early stages(1\&2), which handle high-resolution feature maps, incur higher FLOPs, leading to increased runtime compared to the final stages. Tab.\ref{tab:blocks} illustrates one such analysis: reducing the number of blocks in stage 2 from 6 to 3 results in an 8.2\% reduction in GFLOPs and a 9.6\% decrease in runtime, saving 109 msec. Additionally, the later stages(3\&4) have minimal impact on GFLOPs and runtime since the input resolution decreases for them. This flexibility allows us to adjust the number of blocks in these stages as needed to improve mAP.

\begin{table}
  \caption{Ablation study of number of blocks in the second stage of the backbone network}
  \centering
  \begin{tabular}{ c | c | c | c}
    \hline
    \# blocks & mAP & Model GFLOPs & Model Runtime(m.sec)(\(\Delta \)) \\ \hline
    6 & 27.4 &  18.96 & 1136  \\
    \textbf{3} & \textbf{26.17}  & \textbf{ 17.4} & \textbf{1027(-109)}\\
    \hline
  \end{tabular}
  \label{tab:blocks}
\end{table}

\section{CONCLUSION}

In this work, we successfully achieved deployability of radar object detection on non-GPU-based device Raspberry Pi. This was accomplished by introducing the Feature Enhancement and Compression module to the feature encoder, enabling efficient feature learning and mitigating memory bottlenecks in the early stages and incorporating Depthwise Separable Convolutions into the network backbone, ensuring accuracy is maintained while minimizing the latency. Through extensive investigations of different design choices, we proposed two models: DSFEC-M and DSFEC-S. We have demonstrated the performance of these models in terms of accuracy, model size and speed. Our results suggests that the DSFEC-M model prioritizes efficiency with higher accuracy and decent latency, making it suitable for scenarios where accuracy is paramount. Conversely, the DSFEC-S model is optimized for deployability with low latency and decent accuracy, making it ideal for resource-constrained devices.

\bibliographystyle{unsrt}  
% \bibliography{references}

\end{document}